\documentclass[twoside,11pt]{article}
\usepackage{pgm}

\usepackage[utf8]{inputenc}
\inputencoding{utf8}

\usepackage{hyperref,color,soul,booktabs}
\setulcolor{blue}
\newcommand{\BibTeX}{\textsc{B\kern-0.1emi\kern-0.017emb}\kern-0.15em\TeX}

\ShortHeadings{Differentiable TAN Structure Learning for Bayesian Network Classifiers}{Roth and Pernkopf}

\usepackage{amsmath}
\usepackage{subfig}

\graphicspath{
{./figures/bn_structures/4nodes/}
{./figures/heuristic_structure_mnist/}
{./figures/compare/letter/}
{./figures/compare/satimage/}
{./figures/compare/usps/}
{./figures/compare/mnist/}
{./figures/orderings/letter/}
{./figures/orderings/satimage/}
{./figures/orderings/usps/}
{./figures/orderings/mnist/}
{./figures/parent_subsets/letter/}
{./figures/parent_subsets/satimage/}
{./figures/parent_subsets/usps/}
{./figures/parent_subsets/mnist/}
}

\newcommand{\argmax}{\operatornamewithlimits{argmax}}
\DeclareMathOperator{\softmax}{softmax}
\newcommand{\rvc}{\ensuremath{C}}
\newcommand{\rvx}{\ensuremath{X}}

\newcommand{\rvxset}{\ensuremath{\boldsymbol{X}}}
\newcommand{\vecs}{\mathbf{s}}
\newcommand{\vecx}{\mathbf{x}}

\newcommand{\vecPhi}{\boldsymbol{\Phi}}
\newcommand{\vectheta}{\boldsymbol{\theta}}
\newcommand{\vecTheta}{\boldsymbol{\Theta}}
\newcommand{\vecepsilon}{\boldsymbol{\varepsilon}}
\newcommand{\dataset}{\mathcal{D}}
\newcommand{\graph}{\mathcal{G}}
\newcommand{\parent}{\mathrm{pa}}

\newcommand{\E}{\mathbb{E}}
\DeclareMathOperator{\Unif}{\mathcal{U}}
\DeclareMathOperator{\Gumbel}{Gumbel}

\newcommand{\loss}{\mathcal{L}}
\newcommand{\lossll}{\loss_{\mathrm{NLL}}}
\newcommand{\losslm}{\loss_{\mathrm{LM}}}
\newcommand{\losshyb}{\loss_{\mathrm{HYB}}}
\newcommand{\losssl}{\loss_{\mathrm{SL}}}
\newcommand{\lossgraph}{\loss_{\graph}}

\begin{document}

\title{Differentiable TAN Structure Learning\\for Bayesian Network Classifiers}

\author{\Name{Wolfgang~Roth} \Email{roth@tugraz.at}\and
   \Name{Franz~Pernkopf} \Email{pernkopf@tugraz.at}\\
   \addr Graz University of Technology, Austria\\ Signal Processing and Speech Communication Laboratory}

\maketitle

\begin{abstract}
Learning the structure of Bayesian networks is a difficult combinatorial optimization problem.
In this paper, we consider learning of tree-augmented na\"ive Bayes (TAN) structures for Bayesian network classifiers with discrete input features.
Instead of performing a combinatorial optimization over the space of possible graph structures, the proposed method learns a distribution over graph structures.
After training, we select the most probable structure of this distribution.
This allows for a joint training of the Bayesian network parameters along with its TAN structure using gradient-based optimization.
The proposed method is agnostic to the specific loss and only requires that it is differentiable.
We perform extensive experiments using a hybrid generative-discriminative loss based on the discriminative probabilistic margin.
Our method consistently outperforms random TAN structures and Chow-Liu TAN structures.
\end{abstract}

\section{Introduction} \label{sec:introduction}
Many existing approaches to Bayesian network (BN) structure learning are scored-based approaches that maximize a score over the combinatorial space of BN graphs $\graph$ with respect to a dataset $\dataset$.
Structure learning is a highly non-trivial task since the number of graphs $\graph$ is typically superexponential in the number of variables and score maximization is known to be NP-hard, even for the favorable case of decomposable scores \citep{Chickering2004}.

We consider the learning of BNs with tree-augmented na\"ive Bayes (TAN) structure.
While there exists a polynomial time algorithm to learn TAN structures according to the likelihood score \citep{Friedman1997}, this is not the case for commonly used discriminative criteria.
However, discriminative criteria typically yield superior classification performance, which has led to a bulk of literature investigating TAN structure learning with discriminative scores.
\cite{Grossman2004} and \cite{Pernkopf2011} applied greedy hill climbing to a conditional likelihood score and to a probabilistic margin score, respectively.
\cite{Pernkopf2013} reported improved results for the probabilistic margin when optimized with simulated annealing.
\cite{Peharz2012} performed margin based structure learning using a general purpose branch and bound algorithm.
However, these methods have in common that their discriminative score is based on generative maximum likelihood parameters or, as in \citep{Grossman2004}, they require a time-consuming iterative optimization procedure within the hill climbing loop.

Interestingly, the deep learning community is facing similar challenges when designing the structure of deep neural networks.
Whereas deep neural networks used to perform best when designed by an expert, automatic neural architecture search approaches have recently shown to achieve state-of-the-art performance.
In particular, our work is inspired by differentiable architecture search approaches \citep{Liu2019,Cai2019} which are appealing as they train the parameters and the structure of a deep neural network according to the same discriminative criterion through gradient-based optimization without requiring combinatorial search heuristics.

We propose a differentiable approach for TAN structure learning of BN classifiers.
Our method assumes a fixed variable ordering and is based on a relaxation of the discrete graph structure to a \emph{discrete distribution} over graph structures.
Given a differentiable loss $\loss$ over the BN parameters, we formulate a new structure learning loss $\losssl = \E[\loss]$ as an expected loss with respect to this distribution.
Subsequently, the structure loss $\losssl$ is jointly optimized for the BN parameters and the \emph{continuous} distribution parameters using gradient-based learning.
After learning, we select the most probable structure.
This allows us to apply commonly used discriminative criteria, such as the conditional likelihood or the probabilistic margin \citep{Roth2018}, for structure learning.
In fact, the proposed method is agnostic to the specific loss and only requires that it is differentiable.

The proposed method implicitly utilizes properties from stochastic mini-batch optimization to avoid local minima.
This is a well-known problem of combinatorial search heuristics for which several techniques have been proposed, such as more elaborate search spaces \citep{Teyssier2005} and perturbation methods \citep{Elidan2002}.

We perform extensive experiments using a hybrid generative-discriminative loss based on the probabilistic margin \citep{Roth2018}.
Our method consistently outperforms random TAN structures and likelihood optimized Chow-Liu TAN structures \citep{Friedman1997} on all evaluated datasets by a large margin.
We show that a heuristic variable ordering for image data based on pixel locality further improves performance.
Our method does not require combinatorial optimization heuristics and can be easily implemented using modern automatic differentiation frameworks.\footnote{Code available online at \url{https://github.com/wroth8/bnc}}

An interesting orthogonal approach concerning continuous optimization of graph structures has been recently proposed by \cite{Zheng2018}.
Their work is based on a continuous formulation of the acyclicity constraint for graphs, allowing for gradient-based optimization.

\section{Background} \label{sec:background}
Throughout the paper, uppercase symbols $\rvx$ and $\rvc$ refer to random variables, and lowercase variables $x$ and $c$ refer to concrete instantiations of these variables.
Similarly, $p(\rvx)$ refers to a distribution, whereas $p(x)$ refers to the probability mass of a concrete instantiation.
We denote vectors of random variables and instantiations using boldface symbols $\rvxset$ and $\vecx$, respectively.

\subsection{Bayesian Networks} \label{sec:bayesian_networks}
Let $\rvxset = \{\rvx_1, \ldots, \rvx_D\}$ be a multivariate random variable.
A BN is a graphical representation of a probability distribution $p(\rvxset)$ that defines a factorization of $p(\rvxset)$ via a directed acyclic graph $\graph$ containing $D$ nodes, each corresponding to a random variable $\rvx_i$.
Let $\parent(\rvx_i)$ be the set of parents of $\rvx_i$ in $\graph$.
Then the graph $\graph$ determines the factorization of the joint distribution as
\begin{align}
  p(\rvxset) = \prod_{i=1}^{D} p\left(\rvx_i \ \vert \ \parent(\rvx_i)\right). \label{eq:bn_factorization}
\end{align}
For nodes $\rvx_i$ that do not have any parents in $\graph$, the corresponding factor in \eqref{eq:bn_factorization} is an unconditional distribution $p(\rvx_i)$.
The full joint distribution $p(\rvxset)$ can now be conveniently specified by the individual factors $p(\rvx_i | \parent(\rvx_i))$.
Throughout this paper, we consider distributions over discrete random variables such that each conditional distribution $p(\rvx_i | \parent(\rvx_i))$ can be represented by a conditional probability table (CPT) $\vectheta_{i|\parent(i)}$, where---with slight abuse of notation---$\parent(i)$ refers to the indices of $\rvx_i$'s parents.
The joint distribution \eqref{eq:bn_factorization} is then determined by the parameters $\vectheta_{\graph} = \{\vectheta_{1|\parent(1)}, \ldots, \vectheta_{D|\parent(D)} \}$.

For classification tasks, we are given---in addition to the random variables $\rvxset$ which are now called features---another random variable $\rvc$ that takes the role of a class variable.
We can then employ a BN to perform classification according to the most probable class $c$ conditioned on $\vecx$, i.e.,
\begin{align}
  \argmax_{c} p(c|\vecx) = \argmax_{c} p(\vecx, c). \label{eq:probabilistic_classifier}
\end{align}
Assuming that the CPTs $\vectheta_{\graph}$ are given as log-probabilities, we can exploit the factorization \eqref{eq:bn_factorization} to compute \eqref{eq:probabilistic_classifier} efficiently by accumulating only $C \cdot (D+1)$ values.

However, the size of a CPT $\vectheta_{i|\parent(i)}$ is given by the number of values that $\rvx_i$ and $\parent(\rvx_i)$ can take jointly.
Consequently, assuming that each random variable can take at least two values, the size of $\rvx_i$'s CPT grows exponentially with the number of parents $|\parent(\rvx_i)|$.
Therefore, it is desirable to maintain graph structures where each node has only few parents such that inference tasks remain feasible.
In this paper, we consider two particularly simple BN structures, namely the na\"ive Bayes structure and TAN structures.
These structures restrict the number of conditioning parents and, therefore, do not suffer from the exponential growth.

\begin{figure}[!t]
\centering
\subfloat[Na\"ive Bayes]{\includegraphics[width=0.22\textwidth]{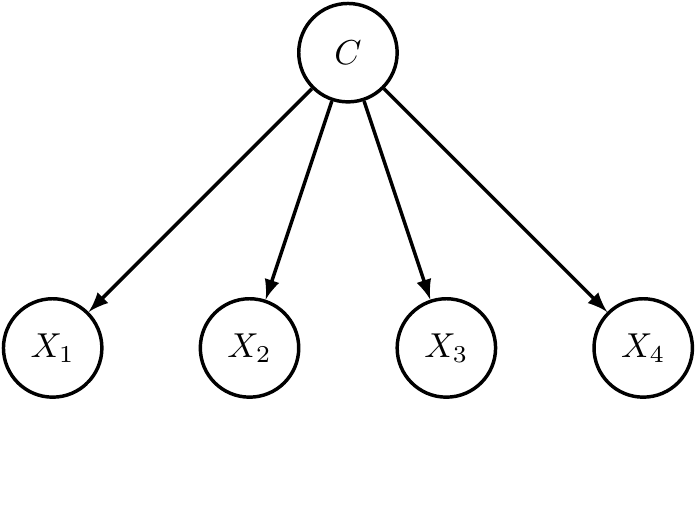}\label{fig:naive_bayes}}
\hfil
\subfloat[TAN (initial)]{\includegraphics[width=0.22\textwidth]{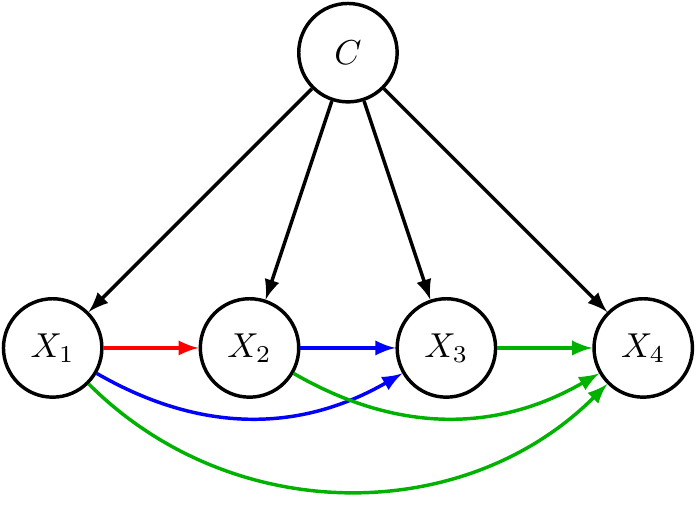}\label{fig:tan_structure_learning_1}}
\hfil
\subfloat[TAN (learned)]{\includegraphics[width=0.22\textwidth]{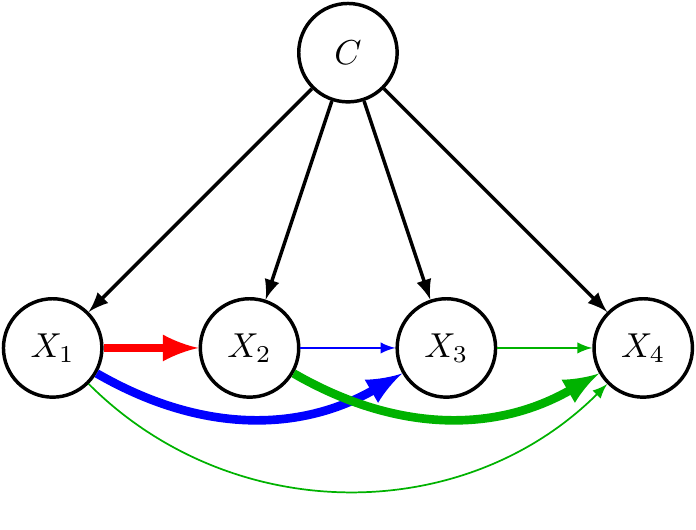}\label{fig:tan_structure_learning_2}}
\hfil
\subfloat[TAN (final)]{\includegraphics[width=0.22\textwidth]{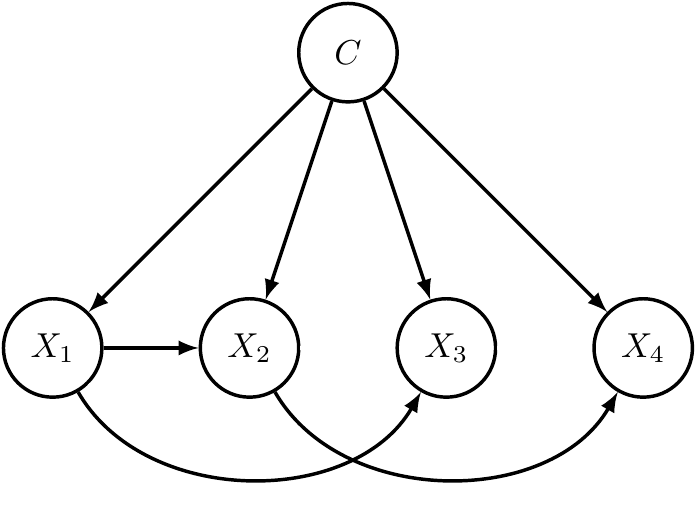}\label{fig:tan_structure_learning_3}}
\caption{%
\protect\subref{fig:naive_bayes} The na\"ive Bayes model as a BN. \protect\subref{fig:tan_structure_learning_1}-\protect\subref{fig:tan_structure_learning_3} TAN structure learning: We consider every left-to-right edge of the ordered variables as a candidate for the conditioning parents.
The TAN structure allows each feature node $\rvx_i$ to have one additional parent besides $\rvc$.
Therefore, each edge is associated with a selection probability indicated by its thickness.
\protect\subref{fig:tan_structure_learning_1} Initially, all edges are equally probable.
\protect\subref{fig:tan_structure_learning_2} After learning, those edges leading to the best objective value are more probable.
\protect\subref{fig:tan_structure_learning_3} The most probable incoming edge of each node is selected as the conditioning parent.}
\label{fig:bayesian_networks}
\end{figure}

\subsection{Na\"ive Bayes and Tree-augmented Na\"ive Bayes (TAN) Structures} \label{sec:nb_tan_structures}
The na\"ive Bayes assumption asserts that all input features $\rvxset$ are conditionally independent given the class variable $\rvc$.
In terms of a BN, this can be modeled as a graph $\graph$ with a single root node $\rvc$ and all variable nodes $\rvx_i$ having $\rvc$ as their sole parent.
This is illustrated in \figurename~\ref{fig:naive_bayes}.
The factorization induced by the na\"ive Bayes assumption is given by
\begin{align}
  p(\rvxset, \rvc) = p(\rvc) \prod_{i=1}^{D} p\left(\rvx_i \ \vert \ \rvc \right).
\end{align}
Although this independence assumption rarely holds in practice, na\"ive Bayes models often perform reasonably well while requiring only few parameters and allowing for fast inference.

To extend the expressiveness of na\"ive Bayes models, TAN models allow each feature $\rvx_i$---in addition to the class variable $\rvc$---to directly depend on one additional feature $\rvx_j$.
This is illustrated in \figurename~\ref{fig:tan_structure_learning_3}.
Consequently, the factorization of a TAN BN is given by
\begin{align}
  p(\rvxset, \rvc) = p(\rvc) \prod_{i=1}^{D} p\left(\rvx_i \ \vert \ \parent(\rvx_i) \right),
\end{align}
subject to the constraints $|\parent(\rvx_i)| \leq 2$ and $\rvc \in \parent(\rvx_i)$.
This slight relaxation of the graph structure is often sufficient to substantially improve the predictive performance compared to na\"ive Bayes models (see Section \ref{sec:experiments}).
However, as opposed to the na\"ive Bayes assumption where the corresponding graph $\graph$ is fixed, selecting a good TAN structure is non-trivial since the space of TAN graphs grows exponentially in the number of nodes $D$.

Nevertheless, there exists a polynomial time algorithm to find a maximum likelihood TAN structure \citep{Friedman1997}.
This algorithm is an extension of the algorithm proposed by \cite{Chow1968} to compute maximum likelihood tree structured BNs.
The algorithm first computes a maximum spanning tree in a fully-connected graph whose nodes correspond to $\rvxset$ and whose edge weights are determined by the conditional mutual information $I(\rvx_i; \rvx_j | \rvc)$, followed by transforming the resulting undirected tree into a directed tree.
We refer to structures discovered by this algorithm as Chow-Liu structures.

However, as reported by several works, generative models tend to perform worse on classification tasks than discriminatively learned models. 
This suggests that the Chow-Liu structure might be suboptimal for classification.
In Section \ref{sec:hybrid_gen_disc_training}, we propose a hybrid generative-discriminative training criterion, and in Section \ref{sec:differentiable_structure_learning}, we propose our method to jointly learn TAN structures and the CPTs $\vectheta_{\graph}$ according to this criterion by means of gradient-based optimization.

\subsection{Hybrid Generative-Discriminative Training} \label{sec:hybrid_gen_disc_training}
Given a dataset $\dataset = \{ (\vecx_n, c_n) \}_{n=1}^N$ comprising $N$ samples of input-target pairs, a probabilistic classifier can be trained using a generative or a discriminative approach.
A \emph{generative} approach is concerned with modeling the joint distribution $p(\rvxset, \rvc)$.
This is typically accomplished by minimizing the negative log-likelihood loss
\begin{align}
  \lossll(\vectheta_{\graph}; \dataset) = - \sum_{n=1}^{N} \log p(\vecx_n, c_n).
\end{align}
A \emph{discriminative} approach is concerned with modeling the conditional distribution $p(\rvc | \rvxset)$ directly.
In this work, we consider a discriminative loss based on the notion of a probabilistic margin \citep{Pernkopf2012,Roth2018}.
In particular, we minimize
\begin{align}
  \losslm(\vectheta_{\graph}; \dataset) = \sum_{n=1}^{N} \max(0, \gamma - \beta_n(\vectheta_{\graph})), \label{eq:loss_lm}
\end{align}
where $\gamma > 0$ is a desired log-margin hyperparameter and $\beta_n$ is the probabilistic log-margin of the $n$\textsuperscript{th} sample defined as
\begin{align}
  \beta_n \left(\vectheta_{\graph}\right) = \log \left( \frac{p(c_n | \vecx_n)}{\max_{c \neq c_n} p(c | \vecx_n)} \right) = \log p(\vecx_n, c_n) - \max_{c \neq c_n} \log p(\vecx_n, c). \label{eq:probabilistic_margin}
\end{align}
In our implementation of \eqref{eq:probabilistic_margin}, we employ the softened version of the maximum from \citep{Roth2018}, i.e., $\max(v_1, \ldots, v_L) \approx (\log \sum_{i=1}^L \exp(\eta v_i)) / \eta$, where $\eta > 1$ is a hyperparameter.

In practice, the discriminative approach is often superior in terms of classification performance, but often discards most of the probabilistic semantics of the model.
This motivates a hybrid gen\-er\-a\-tive-discriminative loss to combine the advantages of both approaches as
\begin{align}
  \losshyb(\vectheta_{\graph}; \dataset) = \lossll(\vectheta_{\graph}; \dataset) + \lambda \losslm(\vectheta_{\graph}; \dataset), \label{eq:loss_hybrid}
\end{align}
where $\lambda > 0$ is a hyperparameter. Previous work has shown that by carefully trading off between the generative and the discriminative loss in \eqref{eq:loss_hybrid}, most of the probabilistic semantics can be maintained while achieving good accuracy.
In many cases a hybrid classifier even outperforms a pure discriminative classifier since the generative term can be seen as a regularizer \citep{Peharz2013}.

\section{Differentiable TAN Structure Learning} \label{sec:differentiable_structure_learning}
In this section, we introduce a loss that admits the joint training of the graph $\graph$ and the CPTs $\vectheta_{\graph}$.
We show how this loss can be trained by means of gradient-based optimization using the reparameterization trick \citep{Kingma2014} and the straight-through gradient estimator (STE) \citep{Bengio2013}, both of which are popular methods in the deep learning community.

\subsection{The Structure Learning Loss} \label{sec:structure_loss}
Let $\rvx_1, \ldots, \rvx_D$ be a fixed ordering of the input features.
In our differentiable structure learning approach, for each feature node $\rvx_i$, we consider every $\rvx_j$ with $j < i$ as a candidate for a conditioning parent.
This is illustrated in \figurename~\ref{fig:tan_structure_learning_1}.
Although the search space depends on the particular ordering of $\rvxset$ and does not cover all possible TAN structures, it is convenient as the resulting graph $\graph$ is guaranteed to be acyclic.

To enable a joint-treatment of the graph structure $\graph$ and the CPT parameters, we reformulate $\log p(\rvxset, \rvc)$ of TAN BNs by introducing new parameters $\vecs$ governing the graph structure $\graph$.
In particular, let $\vecs = \{ \vecs_2, \ldots, \vecs_D \}$ where $\vecs_i = (s_{i|1}, \ldots, s_{i|i-1})$ is a one-hot vector such that $s_{i|j} = 1$ iff $\parent( \rvx_i ) = \{ \rvx_j, \rvc \}$.\footnote{This allows us to talk about $\graph$ and $\vecs$ interchangeably.}
Furthermore, let $\vecTheta = \{\vectheta_{\rvc}, \vectheta_1\} \cup \{ \vecTheta_2, \ldots, \vecTheta_D \}$ be the collection of all potential CPTs, where $\vectheta_{\rvc}$ is the class prior, $\vectheta_1$ is the CPT for $\rvx_1$, and $\vecTheta_i = \{ \vectheta_{i|1}, \ldots, \vectheta_{i|i-1} \}$ contains the CPTs for all possible conditioning parents of node $\rvx_i$.
Then the log-joint probability for a given structure $\vecs$ can be expressed as
\begin{align}
  \log p(\rvxset, \rvc) = \log p_{\vectheta_{\rvc}}(\rvc) + \log p_{\vectheta_1}(\rvx_1 | \rvc) + \sum_{i = 2}^{D} \sum_{j = 1}^{i - 1} s_{i|j} \log p_{\vectheta_{i|j}}(\rvx_i | \rvx_j, \rvc), \label{eq:log_joint_structure}
\end{align}
where we made the dependency on the CPTs $\vecTheta$ explicit in the subscript of $p$.
This allows us to generalize an arbitrary log-likelihood based loss $\lossgraph(\vectheta_{\graph})$ for a specific graph $\graph$ to a loss $\loss(\vecTheta, \vecs)$ including the graph structure $\vecs$ and to optimize the structure parameters $\vecs$ and the CPTs $\vecTheta$ jointly.

However, minimizing the combinatorial loss $\loss(\vecTheta, \vecs)$ is problematic as the number of different structures $\vecs$ scales exponentially in the number of features $D$.
To circumvent the combinatorial nature of $\loss(\vecTheta, \vecs)$, we first introduce \emph{continuous} distribution parameters $\vecPhi = (\vecPhi_2, \ldots, \vecPhi_D)$ where $\vecPhi_i = (\phi_{i|1}, \ldots, \phi_{i|i-1})$ and $\sum_{j=1}^{i-1} \phi_{i|j} = 1, \phi_{i|j} \geq 0$.
These parameters $\vecPhi$ induce a probability distribution over the one-hot vectors in $\vecs$ and, consequently, also over the graph structures $\graph$.
We can then express a differentiable structure learning loss $\losssl$ as an expectation with respect to the distribution over graph structures as
\begin{align}
  \losssl(\vecTheta, \vecPhi) = \E_{\vecs \sim p_{\vecPhi}} \left[ \loss(\vecTheta, \vecs) \right] = \E_{\graph \sim p_{\vecPhi}} \left[ \lossgraph(\vectheta_\graph) \right]. \label{eq:structure_loss}
\end{align}
Here $\vectheta_{\graph}$ are the CPT parameters of $\vecTheta$ required by the graph structure $\graph$.
Given an optimal structure $\vecs^{*}$ with respect to $\loss(\vecTheta, \vecs)$, an optimal solution to \eqref{eq:structure_loss} is given by the distribution $\vecPhi^{*}$ putting all mass on the particular structure, i.e., $\vecs^{*}$ itself.
Note that the distribution $\vecPhi$ has no particular interpretation and merely serves for the purpose of obtaining a differentiable loss $\losssl$.

\subsection{Minimizing the Structure Learning Loss} \label{eq:loss_minimization}
The structure loss \eqref{eq:structure_loss} becomes intractable for a moderate number of features $D$, but it can be optimized with stochastic gradient descent methods using Monte-Carlo estimates of the gradient of \eqref{eq:structure_loss}.
The Monte-Carlo estimates of the gradient are obtained via the reparameterization trick \citep{Kingma2014}, which has recently become a popular method for optimizing intractable expectations.
The idea of the reparameterization trick is to sample $\vecs$ by transforming the distribution parameters $\vecPhi$ along with a random sample $\vecepsilon$ drawn from a \emph{fixed parameter-free} distribution $p(\vecepsilon)$ to obtain $\vecs = g(\vecPhi, \vecepsilon)$.
This allows us to compute gradient samples of $\losssl$ with respect to $\vecPhi$ using the backpropagation algorithm.

For a categorical distribution with probabilities $\vecPhi_i$, we can sample a one-hot encoded vector $\vecs_i$ by means of a reparameterization using the Gumbel-max trick \citep{Jang2017,Maddison2017} according to
\begin{align}
  \vecs_i = \argmax_j \ \{ \ \log \phi_{i|j} + \varepsilon_j \ \vert \ j < i \ \}, \label{eq:gumbel_max}
\end{align}
where $\varepsilon_j \sim \Gumbel(0, 1)$, and $\argmax$ computes a \emph{one-hot} encoding of the maximum argument $j$.

However, the gradient of $\argmax(\vecPhi_i)$ is zero almost everywhere and cannot be used for backpropagation.
To overcome this, we employ the STE \citep{Bengio2013}.
The STE approximates the gradient of a zero-derivative function $h$ during backpropagation with the non-zero gradient of a similar function $\tilde{h} \approx h$, i.e.,
\begin{align}
  \frac{\partial \loss}{\partial h} \frac{\partial h}{\partial u} \approx \frac{\partial \loss}{\partial h} \frac{\partial \tilde{h}}{\partial u}, \label{eq:ste}
\end{align}
which allows us to perform gradient-based learning.
In our case, we approximate the gradient of the $\argmax$ in \eqref{eq:gumbel_max} during backpropagation by the gradient of a $\softmax$ function
\begin{align}
  \softmax_j((\log \vecPhi_i + \vecepsilon) / \tau) = \frac{\exp ((\log \phi_{i|j} + \varepsilon_j) / \tau)}{\sum_{j^{\prime}} \exp ((\log \phi_{i|j^{\prime}} + \varepsilon_{j^{\prime}}) / \tau)}, \label{eq:softmax_ste}
\end{align}
where $\tau>0$ is a temperature hyperparameter.
For $\tau \rightarrow \infty$ we obtain a uniform distribution and $\tau \rightarrow 0$ recovers the one-hot encoding.
This particular sampling procedure is also known as straight-through Gumbel softmax approximation \citep{Jang2017}.

In our approach, the number of possible conditioning parents---and, therefore, also the number of CPTs $\vectheta_{i|j}$---is $\mathcal{O}(D^2)$.
Note that although most of the structure parameters $s_{i|j}$ in \eqref{eq:log_joint_structure} equal zero, we still need to consider every conditional log-probability $\log p_{\vectheta_{i|j}}(\rvx_i| \rvx_j, \rvc)$ as they are required for the gradient of $\vecPhi$ using the STE.
Since this is prohibitive for large $D$, we propose to consider for each feature node $\rvx_i$ only a fixed randomly selected parent subset $\{X_j \ \vert \ j < i\}$ of maximum size $K \ll D$.
This results in a linear dependence on the number of features $D$ as $\mathcal{O}(K D)$.

After training has finished, we select the single most probable structure $\graph$ and the corresponding CPTs $\vectheta_{\graph}$.
We emphasize that the presented method for structure learning is agnostic to the particular loss $\lossgraph$ and that it can be used in conjunction with any differentiable loss.

\section{Experiments} \label{sec:experiments}
\subsection{Datasets} \label{sec:datasets}
We conducted experiments on the following datasets also used in \citep{Tschiatschek2014}.
\paragraph{Letter:} This dataset contains 20,000 samples, describing one of 26 English letters using 16 numerical features (statistical moments and edge counts) extracted from images \citep{Dua2019}.
\paragraph{Satimage:} This dataset contains 6,435 samples containing multi-spectral values of $3 \times 3$ pixel neighborhoods in satellite images, resulting in a total of 36 features.
The aim is to classify the central pixel of these image patches to one of the seven categories red soil,  cotton crop, grey soil, damp grey soil, soil with vegetation stubble, mixture class (all types present), very damp grey soil.
\paragraph{USPS:} This dataset contains 11,000 grayscale images of size $16 \times 16$, showing handwritten digits obtained from zip codes of mail envelopes \citep{Hastie2009}.
Every pixel is treated as a feature.
\paragraph{MNIST:} This dataset contains 70,000 grayscale images showing handwritten digits from 0--9 \citep{LeCun1998}.
The original images of size $28 \times 28$ are linearly downscaled to $14 \times 14$ pixels.
Every pixel is treated as a feature.

Except for \emph{satimage}, where we use 5-fold cross-validation, we split each dataset into two thirds of training samples and one third of test samples.
The features were discretized using the approach from \citep{Fayyad1993}.
The average numbers of discrete values per feature are $9.1$, $11.5$, $3.4$, and $13.2$ for the respective datasets in the order presented above.

\subsection{Experiment Setup} \label{sec:experiment_setup}
All experiments were performed using the stochastic optimizer Adam \citep{Kingma2015} for 500 epochs.
We used mini-batches of size $50$ on \emph{satimage}, $100$ on \emph{letter} and \emph{usps}, and $250$ on \emph{mnist}.
Each experiment is performed using the two learning rates $\{3\cdot10^{-3}, 3\cdot10^{-2}\}$ for the CPTs $\vecTheta$, and we report the superior result of the two optimization runs.
The learning rate is decayed exponentially after each epoch, such that it decreases by a factor of $10^{-3}$ over the training run.
We used a fixed learning rate of $10^{-3}$ for the structure parameters $\vecPhi$ that is not decayed.

The CPT parameters $\vecTheta$ and the structure parameters $\vecPhi$ are stored as un\-nor\-mal\-ized log-prob\-a\-bil\-i\-ties.
We initialize $\vecTheta$ randomly using a uniform distribution $\Unif([-0.1, 0.1])$, and we set $\vecPhi$ initially to zero, resulting in a uniform distribution over graphs $\graph$.

We anneal $\tau$ in \eqref{eq:softmax_ste} exponentially from $10^1$ to $10^{-1}$ over the training run.
This results in a more uniform distribution over structures $\vecs$ at the beginning of training, facilitating exploration of different structures, whereas the distribution becomes more concentrated at particular structures towards the end of training.

We compared the classification performance of \textbf{Na\"ive Bayes (NB)} and several TAN structures.
\paragraph{TAN Random:} TAN structure obtained by a random variable ordering and selecting a random parent $\rvx_j$ for each $\rvx_i$ with $j<i$.
We evaluated ten parent selections for five variable orderings, resulting in 50 structures in total.

\paragraph{Chow-Liu:} TAN structure obtained using the procedure from \cite{Friedman1997}.

\paragraph{TAN Subset (ours):} Fixed random variable ordering and randomly selected parent subset of maximum size $K$ satisfying the $j<i$ constraint.
We evaluated five parent subsets for five variable orderings, resulting in 25 configurations in total.
We selected $K \in \{2, 5, 8\}$.

\paragraph{TAN All (ours):} Fixed random variable ordering and considering \emph{all} parents satisfying $j<i$.
We evaluated five variable orderings.
This setting is only evaluated for the datasets \emph{letter} and \emph{satimage} having fewer features $D$.

\paragraph{TAN Heuristic (ours):} Heuristically determined feature ordering and respective parent subsets based on pixel locality.
This setting is evaluated for the image datasets \emph{usps} and \emph{mnist}.
We evaluated three ordering heuristics (for details see Section~\ref{sec:heuristic_structure}) and selected $K \in \{1, 2, 5, 8\}$.

We evaluated the same five random variable orderings for TAN Random, TAN Subset, and TAN All.
We introduced additional probability parameters $\phi_{i|\emptyset}$ and corresponding CPTs $\vectheta_{i|\emptyset}$, allowing $X_i$ to also have no additional parent besides $\rvc$, such that for TAN Subset and TAN Heuristic there are effectively up to $K+1$ choices for the conditioning parents of $\rvx_i$.
To assess the impact of different $K$, the parent subsets for smaller $K$ are strict subsets of parent subsets for larger $K$.

We tuned the hyperparameters of $\losshyb$ using random search in two different settings.
In setting (I), we selected 500 hyperparameter configurations according to $\log_{10} \lambda \sim \Unif([0, 3])$, $\log_{10} \gamma \sim \Unif([-1, 2])$, and $\eta \sim \Unif([1, 20])$.
In setting (II), we selected 100 hyperparameter configurations according to $\log_{10} \lambda \sim \Unif([1, 3])$, $\log_{10} \gamma \sim \Unif([-1, 2])$, and used a fixed $\eta=10$.
We applied (II) to experiments where we evaluated more structural settings, namely TAN Random, TAN Subset, and TAN Heuristic, and we applied setting (I) to all other experiments.

\subsection{Classification Results} \label{sec:classification_results}
\begin{figure}[!t]
\centering
\subfloat[letter]{\includegraphics[scale=0.4825]{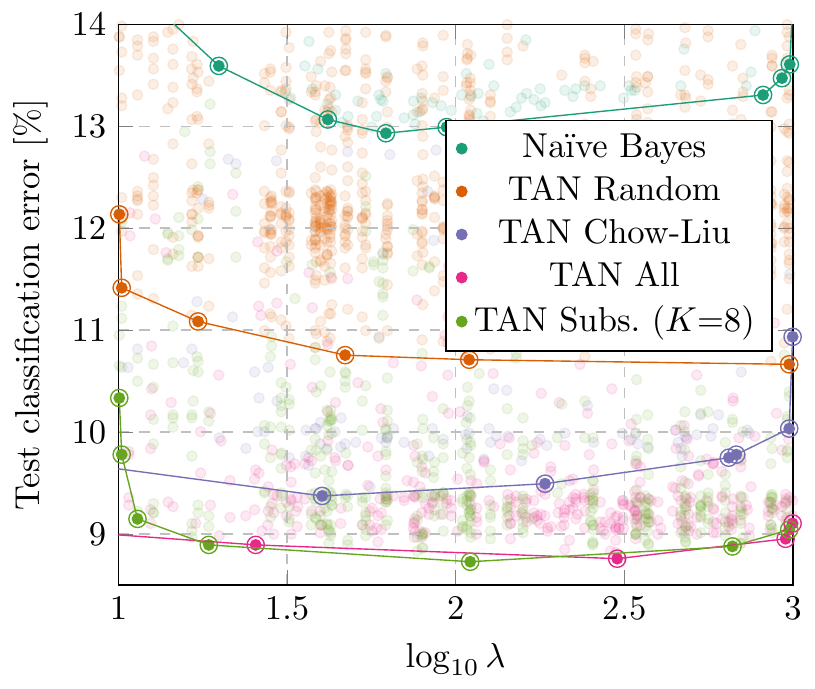}\label{fig:compare_letter}}
\hfil
\subfloat[satimage]{\includegraphics[scale=0.4825]{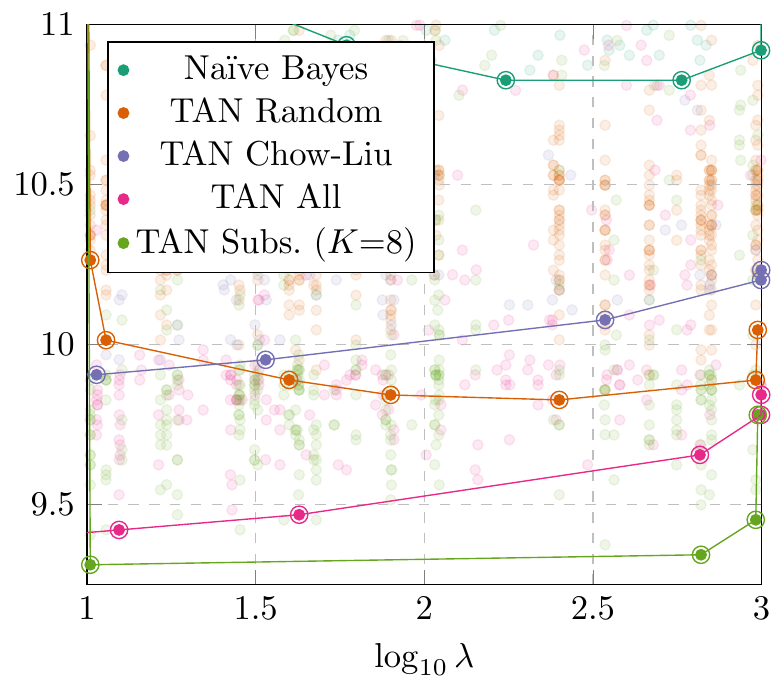}\label{fig:compare_satimage}}
\hfil
\subfloat[usps]{\includegraphics[scale=0.4825]{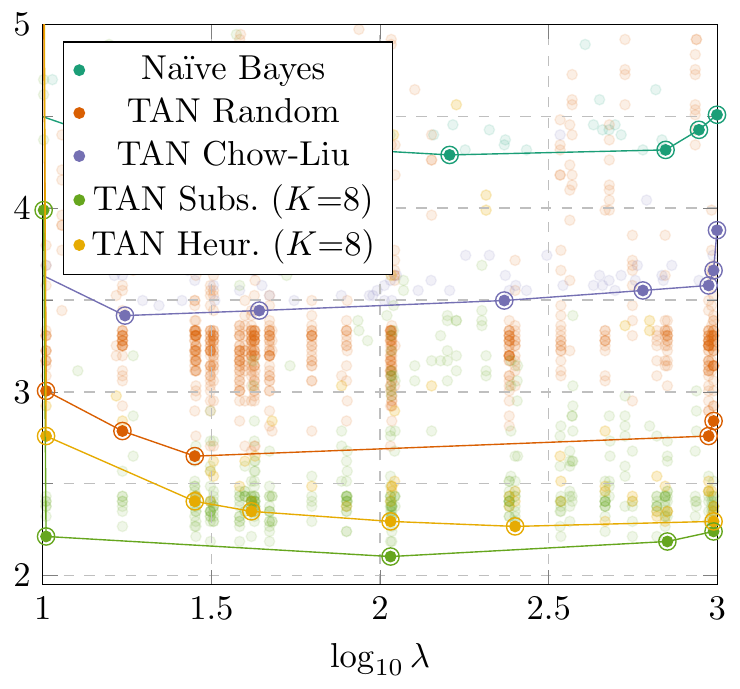}\label{fig:compare_usps}}
\hfil
\subfloat[mnist]{\includegraphics[scale=0.4825]{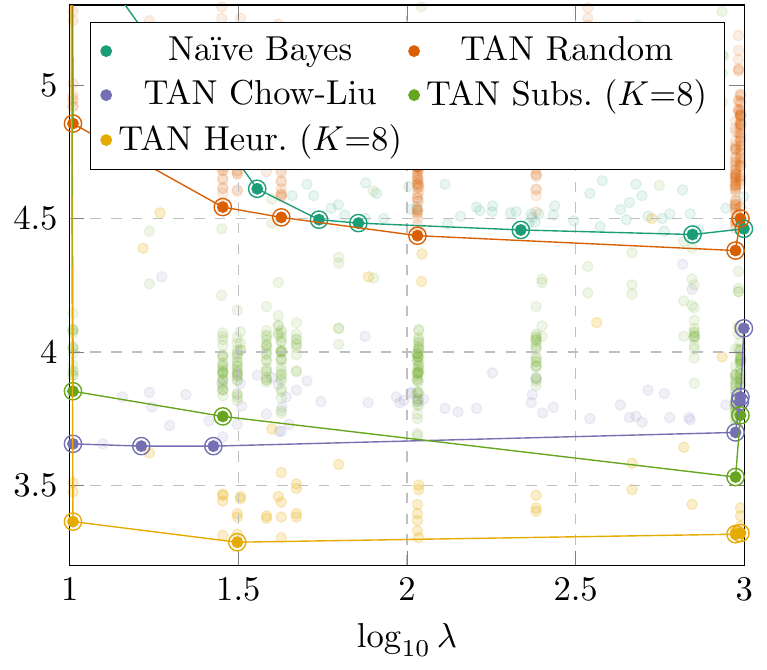}\label{fig:compare_mnist}}
\caption{%
Test classification errors [\%] of different methods over $\log_{10} \lambda$.
Each point corresponds to a different experiment.
For better visualization, the lines show the lower part of the convex hull of equally colored points, i.e., no points of this color are below this line.
}
\label{fig:compare}
\end{figure}

We report the test errors \emph{after} 500 epochs of training.
Results for maximum likelihood parameters (ML) are obtained in closed-form.
The best classification errors [\%] over all parameter settings are shown in \tablename~\ref{tab:results}.
Results of individual experiments are shown in \figurename~\ref{fig:compare}.

Models with generative parameters (ML) perform poorly, showing that discriminative training is beneficial.
The TAN structures outperform Na\"ive Bayes by a large margin, which highlights the benefit of introducing simple TAN interactions, even when they are selected randomly (TAN Random).
Note that the data-driven Chow-Liu structure does not outperform TAN Random on all datasets, and there is even a large performance gap on \emph{usps} where we observed overfitting.

Our method (TAN Subset) outperforms TAN Random and Chow-Liu on all datasets.
We emphasize that our method outperforms these baseline models on a wide range of settings (cf.\ \figurename~\ref{fig:compare}), and not just using the best setting.
TAN Subset achieved its best performance using the largest parent subsets with $K=8$ on all datasets.
Interestingly, TAN All does not benefit when considering \emph{all} conditioning parents compared to TAN Subset with $K=8$.
We attribute this to the fact that \emph{letter} and \emph{satimage} have relatively small numbers of features $D=16$ and $D=36$, respectively, and $K=8$ suffices to cover a good structure with the randomly selected parent subsets with high probability.
However, note that using a larger $K$ reduces the gradient updates per CPT since only the gradient of a single CPT $\vectheta_{i|j}$ for each $i$ is non-zero, resulting in different learning dynamics.
Consequently, a different choice of number of epochs or learning rate might yield improved results.

\begin{table}
\scriptsize
\centering
\begin{tabular}{l|cc|ccc|ccc}
\toprule
loss $\loss$  & \multicolumn{2}{c|}{$\lossll$ (ML)}  & \multicolumn{3}{c|}{$\losshyb$} & \multicolumn{3}{c}{$\losssl$ with $\losshyb$ (ours)} \\
\midrule
structure  & NB  & Chow-Liu & NB & TAN Random & Chow-Liu & TAN Subset & TAN All & TAN Heuristic  \\
\midrule
letter   & $25.89$  & $15.23$  & $12.93$       & $10.66$    &  $9.37$      &  $8.73$           &  $8.76$        & / \\
satimage & $17.95$  & $11.90$  & $10.83$       &  $9.83$    &  $9.91$      &  $9.31$           &  $9.39$        & / \\
usps     & $13.11$  &  $8.74$  &  $4.29$       &  $2.65$    &  $3.42$      &  $2.10$           & /              &  $2.27$ \\
mnist    & $17.34$  &  $7.03$  &  $4.44$       &  $4.38$    &  $3.65$      &  $3.53$           & /              &  $3.29$ \\
\bottomrule
\end{tabular}
\caption{Classification errors [\%] of various BN structures on several datasets. See text for details.}\label{tab:results}
\end{table}

\subsection{Heuristic Structures for Image Data} \label{sec:heuristic_structure}
We evaluated three different heuristic feature orderings of quadratically sized images.
Feature ordering A considers the pixels row-wise from top to bottom in a left-to-right fashion.
Feature ordering B proceeds along the main diagonal by traversing the lower triangular image matrix (including the diagonal) in a row-wise fashion, and including after each pixel the corresponding transposed pixel from the upper triangular matrix.
Feature ordering C proceeds from the center of the image outwards.
Assuming that a center region of $H \times H$ is already ordered, we add the $H$ pixels directly above and the $H$ pixels directly below in a left-to-right fashion, and then we add the $H$ pixels directly to the left and the $H$ pixels directly to the right in a top-to-bottom fashion.
Finally, we also include the four corner pixels in the order left-above, right-above, left-below, and right-below.
For each ordering A, B, and C, the subset of at most $K$ conditioning parents of $\rvx_i$ is obtained by selecting $\{X_j \vert j < i\}$ as the $K$ closest features with respect to Euclidean distance of the corresponding pixel locations.

\begin{figure}[!t]
\centering
\subfloat[mnist]{\includegraphics[scale=0.4825]{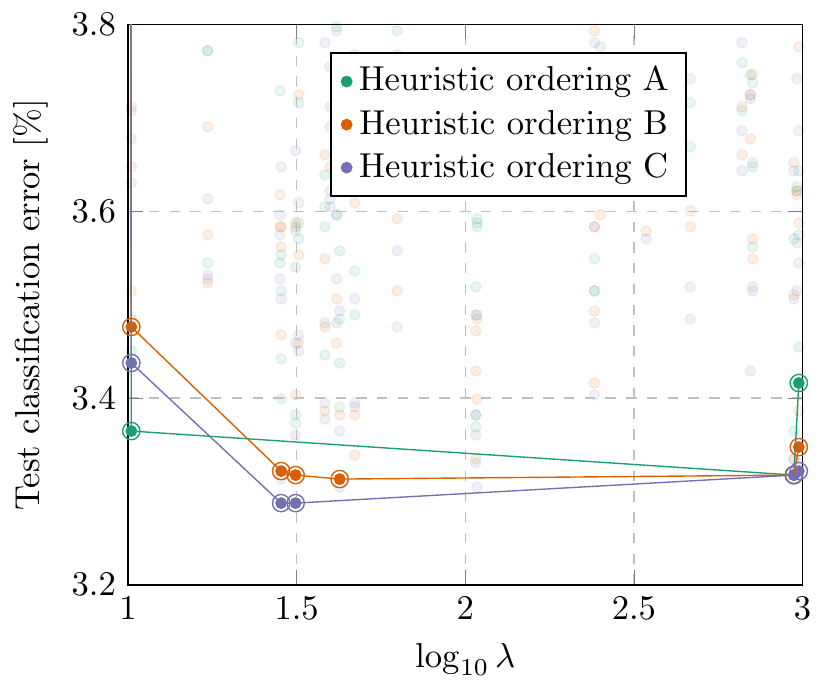}\label{fig:heuristic_cmp_ordering}}
\hfil
\subfloat[mnist]{\includegraphics[scale=0.4825]{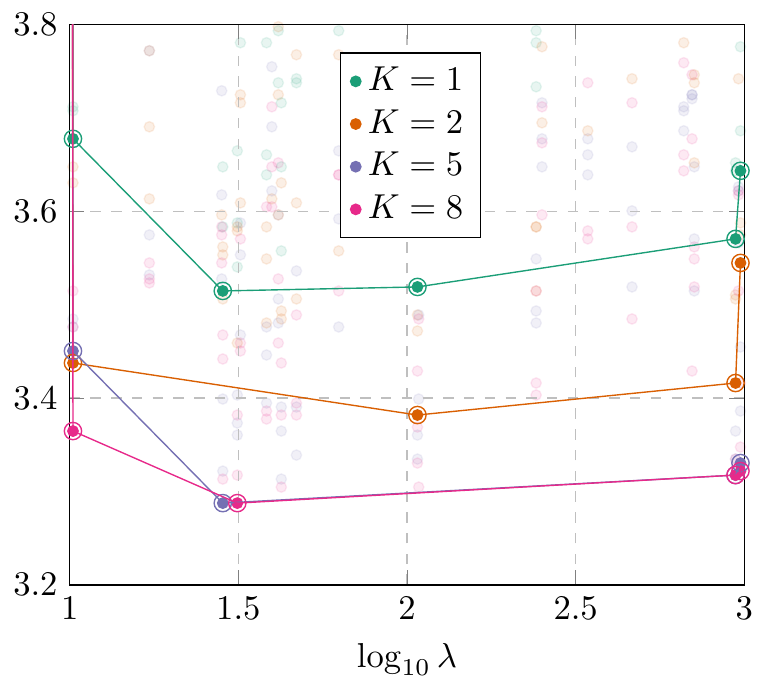}\label{fig:heuristic_cmp_K}}
\hfil
\subfloat[satimage]{\includegraphics[scale=0.4825]{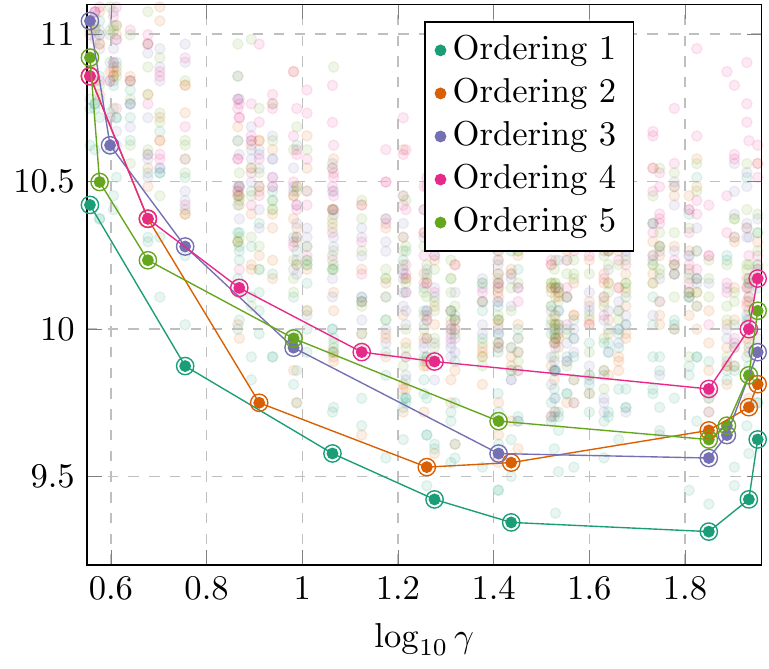}\label{fig:ordering_satimage}}
\hfil
\subfloat[letter]{\includegraphics[scale=0.4825]{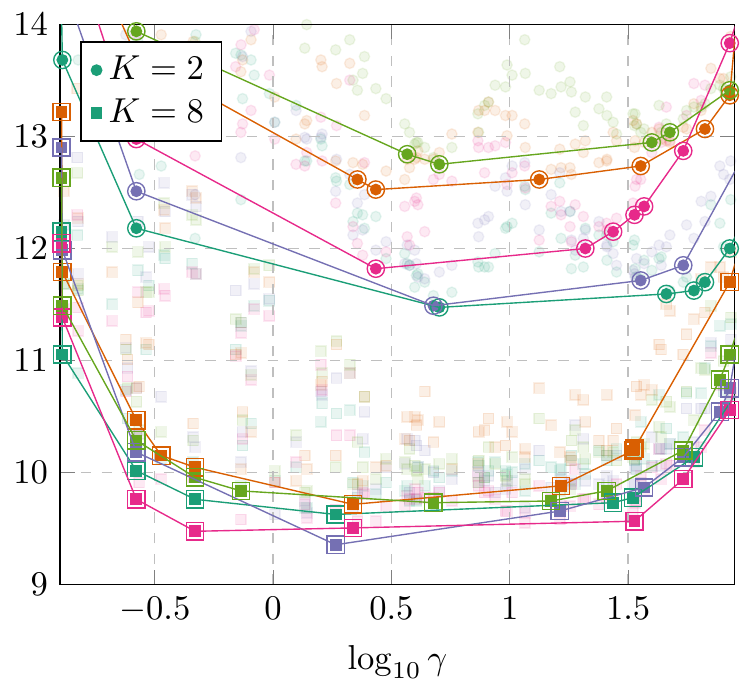}\label{fig:conditioning_parents_letter}}
\caption{%
Test classification errors [\%] over a hyperparameter.
Each point corresponds to a different experiment.
For better visualization, the lines show the lower part of the convex hull of equally colored points, i.e., no points of this color are below this line.
\protect\subref{fig:heuristic_cmp_ordering} Results of TAN Heuristic on \emph{mnist} with an emphasis on different heuristic feature orderings.
\protect\subref{fig:heuristic_cmp_K} Same results as in \protect\subref{fig:heuristic_cmp_ordering}, but with an emphasis on the maximum number of conditioning parents $K \in \{1,2,5,8\}$.
\protect\subref{fig:ordering_satimage} Results of TAN Subset on \emph{satimage} for several feature orderings.
\protect\subref{fig:conditioning_parents_letter} Results of TAN Subset on \emph{letter} for several parent subsets (encoded as colors) and $K \in \{2, 8\}$ evaluated for a fixed feature ordering.
}
\label{fig:results_heuristic}
\end{figure}

The heuristic structures show clear improvements on \emph{mnist} (see \tablename~\ref{tab:results} and \figurename~\ref{fig:compare_mnist}), and the best result was obtained using ordering C.
Overall, when comparing different runs over several random hyperparameter settings in \figurename~\ref{fig:heuristic_cmp_ordering}, we found that ordering C slightly outperforms ordering B which in turn slightly outperforms ordering A.
Note that any of the proposed heuristic orderings outperform the random orderings of TAN Subset.

\figurename~\ref{fig:heuristic_cmp_K} compares distinct values of $K$.
Overall, using larger parent subsets improves the performance.
When comparing $K=1$ and $K=2$, we can see that choosing between two neighboring pixels and not just selecting a fixed one improves performance.
When going from $K=5$ to $K=8$, the accuracy gains are marginal, showing that our method works well for smaller $K$ if the feature ordering and parent subsets are carefully selected.
We emphasize that this latter behavior is specific to the heuristic structures, and we generally observed larger gains when going from $K=5$ to $K=8$ for TAN Subset.

We did not observe benefits of the heuristic structures on \emph{usps}, which we attribute to overfitting issues similar as to why Chow-Liu performs worse than TAN Random on this dataset.

\subsection{Influence of the Feature Ordering and Parent Subsets}
The influence of the feature ordering of TAN Subset on \emph{satimage} for $K=8$ is shown in \figurename~\ref{fig:ordering_satimage}.
We can see that some orderings clearly outperform others over a wide range of parameters, showing that the selected ordering might have a large impact on the overall performance.

To further assess the influence of fixing the feature ordering, we conducted an experiment by allowing conditioning parents that violate the $j < i$ constraint.
This results in pseudo TAN structure classifiers which potentially contain cycles and, therefore, are not BNs anymore.
The classification errors remained similar on all datasets except \emph{letter} where we achieved $7.98\%$ ($0.75\%$ absolute improvement).
These findings suggest that more elaborate techniques considering different orderings as in \citet{Zheng2018} are worthy of investigation.

Next, we investigated the influence of particular parent subsets.
Figure~\ref{fig:conditioning_parents_letter} shows different parent subsets of TAN Subset on \emph{letter} for $K \in \{2, 8\}$.
Again, using larger $K=8$ clearly outperforms the smaller $K=2$.
Similar to feature orderings, some parent subsets clearly outperform others over a wide range of parameters, but here the gap can be reduced by increasing $K$.

\subsection{Recovering the Chow-Liu Structure}
We conducted an experiment using the generative loss $\lossll$ ($\lambda=0$) to see whether our approach is capable of recovering the ``ground truth'' Chow-Liu structure.
Therefore, we computed a Chow-Liu structure and a corresponding ordering such that the Chow-Liu structure is contained in the search space.
Our method was able to recover the Chow-Liu structure consistently.
Note that this is a minimal requirement of our structure learning approach as the structure learning loss $\losssl$ decomposes to local terms similar as the generative loss $\lossll$, substantially simplifying the optimization problem.

\section{Conclusion} \label{sec:conclusion}
We have presented an approach to jointly train the parameters of a BN along with its graph structure through gradient-based optimization.
The method can be easily implemented using modern automatic differentiation frameworks and does not require combinatorial optimization techniques, such as greedy hill climbing.
The presented method is agnostic to the specific loss and can be used with any differentiable loss.
We have presented extensive experiments using a hybrid generative-discriminative loss based on the probabilistic margin, showing that our method consistently outperforms random TAN structures and Chow-Liu TAN structures.
Our method can be combined with heuristic variable orderings to further improve performance.

The performance of our method depends on the choice of a fixed feature ordering and (for TAN Subset) on the choice of a particular parent subset of maximum size $K$.
Consequently, increasing scalability to larger $K$ and investigating methods not requiring a fixed ordering such as in \citet{Zheng2018} are interesting directions of future research.
Furthermore, we believe that the underlying principles of our approach are not restricted to TAN structures and that our approach can be employed to learn other classes of (BN) graph structures.

\appendix
\acks{This work was supported by the Austrian Science Fund (FWF) under the project number I2706-N31.}
{\small
\bibliography{roth20a}

\begin{thebibliography}{25}
\providecommand{\natexlab}[1]{#1}
\providecommand{\url}[1]{\texttt{#1}}
\expandafter\ifx\csname urlstyle\endcsname\relax
  \providecommand{\doi}[1]{doi: #1}\else
  \providecommand{\doi}{doi: \begingroup \urlstyle{rm}\Url}\fi

\bibitem[Bengio et~al.(2013)Bengio, L{\'{e}}onard, and Courville]{Bengio2013}
Y.~Bengio, N.~L{\'{e}}onard, and A.~C. Courville.
\newblock Estimating or propagating gradients through stochastic neurons for
  conditional computation.
\newblock \emph{CoRR}, abs/1308.3432, 2013.

\bibitem[Cai et~al.(2019)Cai, Zhu, and Han]{Cai2019}
H.~Cai, L.~Zhu, and S.~Han.
\newblock {ProxylessNAS}: Direct neural architecture search on target task and
  hardware.
\newblock In \emph{International Conference on Learning Representations
  (ICLR)}, 2019.

\bibitem[Chickering et~al.(2004)Chickering, Heckerman, and
  Meek]{Chickering2004}
D.~M. Chickering, D.~Heckerman, and C.~Meek.
\newblock Large-sample learning of {B}ayesian networks is {NP}-hard.
\newblock \emph{Journal of Machine Learning Research (JMLR)}, 5:\penalty0
  1287--1330, 2004.

\bibitem[Chow and Liu(1968)]{Chow1968}
C.~K. Chow and C.~N. Liu.
\newblock Approximating discrete probability distributions with dependence
  trees.
\newblock \emph{{IEEE} Transactions on Information Theory}, 14\penalty0
  (3):\penalty0 462--467, 1968.

\bibitem[Dua and Graff(2019)]{Dua2019}
D.~Dua and C.~Graff.
\newblock {UCI} machine learning repository, 2019.
\newblock URL \url{http://archive.ics.uci.edu/ml}.

\bibitem[Elidan et~al.(2002)Elidan, Ninio, Friedman, and
  Schuurmans]{Elidan2002}
G.~Elidan, M.~Ninio, N.~Friedman, and D.~Schuurmans.
\newblock Data perturbation for escaping local maxima in learning.
\newblock In \emph{National Conference on Artificial Intelligence (AAAI)},
  pages 132--139, 2002.

\bibitem[Fayyad and Irani(1993)]{Fayyad1993}
U.~Fayyad and K.~Irani.
\newblock Multi-interval discretization of continuous-valued attributes for
  classification learning.
\newblock In \emph{International Joint Conference on Artificial Intelligence
  (IJCAI)}, pages 1022--1027, 1993.

\bibitem[Friedman et~al.(1997)Friedman, Geiger, and Goldszmidt]{Friedman1997}
N.~Friedman, D.~Geiger, and M.~Goldszmidt.
\newblock {B}ayesian network classifiers.
\newblock \emph{Machine Learning}, 29\penalty0 (2--3):\penalty0 131--163, 1997.

\bibitem[Grossman and Domingos(2004)]{Grossman2004}
D.~Grossman and P.~M. Domingos.
\newblock Learning {B}ayesian network classifiers by maximizing conditional
  likelihood.
\newblock In \emph{International Conference on Machine Learning (ICML)},
  volume~69, 2004.

\bibitem[Hastie et~al.(2009)Hastie, Tibshirani, and Friedman]{Hastie2009}
T.~Hastie, R.~Tibshirani, and J.~H. Friedman.
\newblock \emph{The Elements of Statistical Learning: Data Mining, Inference,
  and Prediction, 2nd Edition}.
\newblock Springer Series in Statistics. Springer, 2009.

\bibitem[Jang et~al.(2017)Jang, Gu, and Poole]{Jang2017}
E.~Jang, S.~Gu, and B.~Poole.
\newblock Categorical reparameterization with {G}umbel-softmax.
\newblock In \emph{International Conference on Learning Representations
  (ICLR)}, 2017.

\bibitem[Kingma and Ba(2015)]{Kingma2015}
D.~Kingma and J.~Ba.
\newblock Adam: A method for stochastic optimization.
\newblock In \emph{International Conference on Learning Representations
  (ICLR)}, 2015.

\bibitem[Kingma and Welling(2014)]{Kingma2014}
D.~P. Kingma and M.~Welling.
\newblock Auto-encoding variational {B}ayes.
\newblock In \emph{International Conference on Learning Representations
  (ICLR)}, 2014.

\bibitem[LeCun et~al.(1998)LeCun, Bottou, Bengio, and Haffner]{LeCun1998}
Y.~LeCun, L.~Bottou, Y.~Bengio, and P.~Haffner.
\newblock Gradient-based learning applied to document recognition.
\newblock \emph{Proceedings of the IEEE}, 86\penalty0 (11):\penalty0
  2278--2324, 1998.

\bibitem[Liu et~al.(2019)Liu, Simonyan, and Yang]{Liu2019}
H.~Liu, K.~Simonyan, and Y.~Yang.
\newblock {DARTS}: Differentiable architecture search.
\newblock In \emph{International Conference on Learning Representations
  (ICLR)}, 2019.

\bibitem[Maddison et~al.(2017)Maddison, Mnih, and Teh]{Maddison2017}
C.~J. Maddison, A.~Mnih, and Y.~W. Teh.
\newblock The {C}oncrete distribution: {A} continuous relaxation of discrete
  random variables.
\newblock In \emph{International Conference on Learning Representations
  (ICLR)}, 2017.

\bibitem[Peharz and Pernkopf(2012)]{Peharz2012}
R.~Peharz and F.~Pernkopf.
\newblock Exact maximum margin structure learning of {B}ayesian networks.
\newblock In \emph{International Conference on Machine Learning (ICML)}, 2012.

\bibitem[Peharz et~al.(2013)Peharz, Tschiatschek, and Pernkopf]{Peharz2013}
R.~Peharz, S.~Tschiatschek, and F.~Pernkopf.
\newblock The most generative maximum margin {B}ayesian networks.
\newblock In \emph{International Conference on Machine Learning (ICML)},
  volume~28, pages 235--243, 2013.

\bibitem[Pernkopf and Wohlmayr(2013)]{Pernkopf2013}
F.~Pernkopf and M.~Wohlmayr.
\newblock Stochastic margin-based structure learning of {B}ayesian network
  classifiers.
\newblock \emph{Pattern Recognition}, 46\penalty0 (2):\penalty0 464--471, 2013.

\bibitem[Pernkopf et~al.(2011)Pernkopf, Wohlmayr, and
  M{\"{u}}cke]{Pernkopf2011}
F.~Pernkopf, M.~Wohlmayr, and M.~M{\"{u}}cke.
\newblock Maximum margin structure learning of {B}ayesian network classifiers.
\newblock In \emph{{IEEE} International Conference on Acoustics, Speech, and
  Signal Processing (ICASSP)}, pages 2076--2079, 2011.

\bibitem[Pernkopf et~al.(2012)Pernkopf, Wohlmayr, and
  Tschiatschek]{Pernkopf2012}
F.~Pernkopf, M.~Wohlmayr, and S.~Tschiatschek.
\newblock Maximum margin {B}ayesian network classifiers.
\newblock \emph{{IEEE} Transactions on Pattern Analysis and Machine
  Intelligence}, 34\penalty0 (3):\penalty0 521--532, 2012.

\bibitem[Roth et~al.(2018)Roth, Peharz, Tschiatschek, and Pernkopf]{Roth2018}
W.~Roth, R.~Peharz, S.~Tschiatschek, and F.~Pernkopf.
\newblock Hybrid generative-discriminative training of {G}aussian mixture
  models.
\newblock \emph{Pattern Recognition Letters}, 112:\penalty0 131--137, 2018.

\bibitem[Teyssier and Koller(2005)]{Teyssier2005}
M.~Teyssier and D.~Koller.
\newblock Ordering-based search: {A} simple and effective algorithm for
  learning {B}ayesian networks.
\newblock In \emph{Conference on Uncertainty in Artificial Intelligence (UAI)},
  pages 548--549, 2005.

\bibitem[Tschiatschek et~al.(2014)Tschiatschek, Paul, and
  Pernkopf]{Tschiatschek2014}
S.~Tschiatschek, K.~Paul, and F.~Pernkopf.
\newblock Integer {B}ayesian network classifiers.
\newblock In \emph{European Conference on Machine Learning (ECML)}, pages
  209--224, 2014.

\bibitem[Zheng et~al.(2018)Zheng, Aragam, Ravikumar, and Xing]{Zheng2018}
X.~Zheng, B.~Aragam, P.~Ravikumar, and E.~P. Xing.
\newblock {DAGs} with {NO} {TEARS}: Continuous optimization for structure
  learning.
\newblock In \emph{Advances in Neural Information Processing Systems
  (NeurIPS)}, pages 9492--9503, 2018.

\end{thebibliography}
}
\end{document}